\documentclass[letterpaper,10pt,conference]{ieeeconf}
\IEEEoverridecommandlockouts
\usepackage[T1]{fontenc}
\usepackage{mathptmx}

\usepackage{amsmath,amssymb}
\usepackage{graphicx}
\usepackage{booktabs}
\usepackage{multirow}
\usepackage{cite}
\usepackage{url}
\usepackage{balance}
\usepackage[table]{xcolor}
\usepackage{hyperref}
\definecolor{aeroteal}{HTML}{087E8B}
\definecolor{aerolight}{HTML}{EAF5F4}

\title{\LARGE \bf
AeroIntent: A Synchronized Multimodal Benchmark for UAV Mission Intent Recognition
}
\author{Yutong Chen$^{1}$, Jingxin Mao$^{2}$, Zhiwei Wei$^{1}$, Bing Li$^{1}$, Jianzhong Ni$^{3}$, and Rongqing Zhang$^{2}$%
\thanks{$^{1}$Y. Chen, Z. Wei, and B. Li are with Tongji University.\newline
\texttt{\{2534128,2311769,lizi\}@tongji.edu.cn}}%
\thanks{$^{2}$J. Mao and R. Zhang are with The Hong Kong University of Science and Technology (Guangzhou).\newline
\texttt{jmao913@connect.hkust-gz.edu.cn},\newline
\texttt{rongqingz@tongji.edu.cn}}%
\thanks{$^{3}$J. Ni is with Nanchang Highly Electrical Appliances Co., Ltd.\newline
\texttt{nijz@shec.com.cn}}}
\hypersetup{
  pdftitle={AeroIntent: A Synchronized Multimodal Benchmark for UAV Mission Intent Recognition},
  pdfauthor={Yutong Chen, Jingxin Mao, Zhiwei Wei, Bing Li, Jianzhong Ni, Rongqing Zhang}
}

\begin{document}
\maketitle
\thispagestyle{empty}
\pagestyle{empty}

\begin{abstract}
Reliable recognition of mission-level UAV intent from multimodal observations is essential for low-altitude airspace surveillance and security. Existing multimodal UAV datasets primarily support detection, tracking, and localization, leaving limited support for evaluating mission-intent recognition across held-out trajectory programs and incomplete radar histories. We introduce AeroIntent, a simulation-based benchmark comprising 990 synchronized multimodal sequences generated by 22 trajectory programs across five mission intents. The benchmark provides radar, RGB, infrared, and depth observations, together with program-disjoint, radar-gap, and observation-prefix evaluation protocols. Building on this benchmark, we propose \textbf{AeroMIR}, a structured Multimodal Intent Reasoner that integrates radar trajectories, derived behavioral features, selected visual evidence, and task knowledge for candidate-based UAV intent recognition. Under a strict program-disjoint protocol evaluating unseen tactical realizations, AeroMIR achieves a mean accuracy of \textbf{71.56\%} and a mean Macro-F1 of \textbf{69.84\%} across three seeds, outperforming the strongest trajectory-only baseline by \textbf{7.79} Macro-F1 points and a conventional radar--RGB late-fusion baseline by \textbf{19.32} points. AeroMIR also demonstrates strong performance under controlled radar gaps and early intent recognition settings. AeroIntent and AeroMIR provide a reproducible foundation for multimodal UAV intent recognition. The benchmark is available at: \href{https://anonymous.4open.science/r/AeroIntent}{https://anonymous.4open.science/r/AeroIntent}.

\end{abstract}

\suppressfloats[t]
\begin{figure}[t]
    \centering
    \includegraphics[width=\columnwidth]{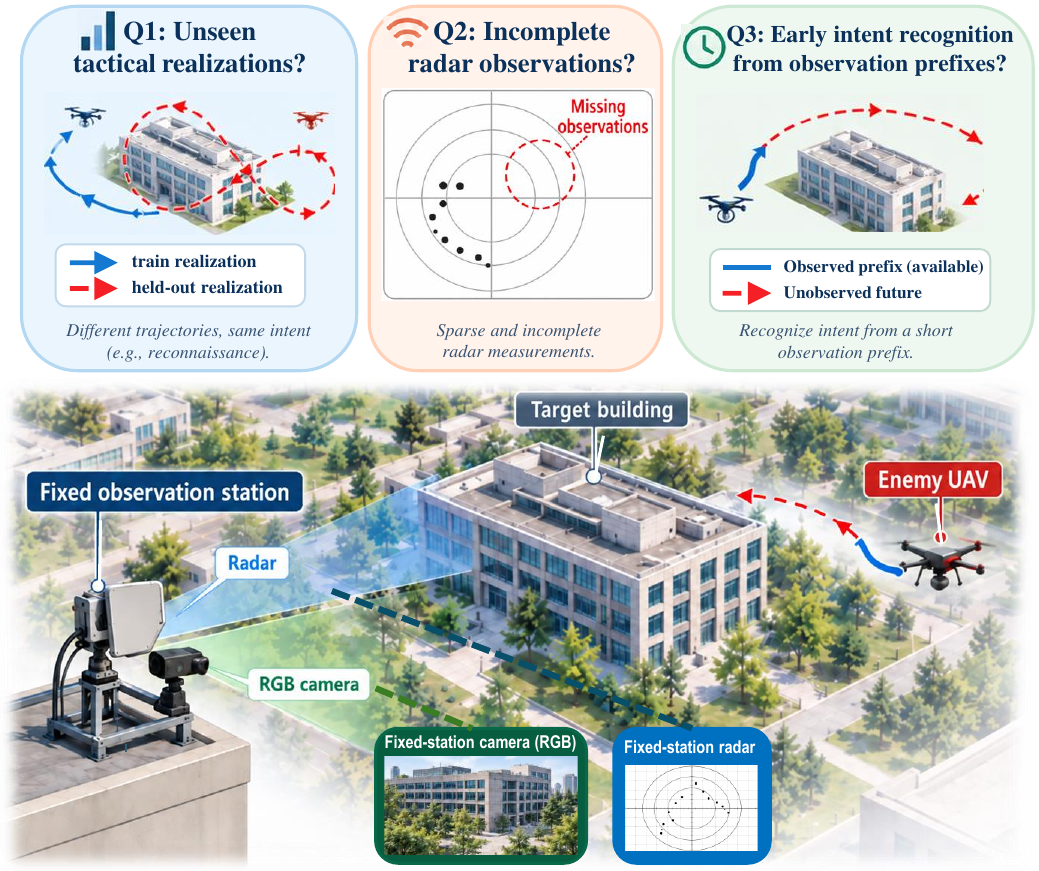}
\caption{\textbf{Observation scenario and key challenges of UAV intent recognition.}
A fixed observation station combines radar measurements and RGB imagery to infer the intent of a UAV near a monitored building. The three research questions concern generalization to unseen trajectories expressing the same intent (Q1), robustness to incomplete radar observations (Q2), and early recognition from short observation prefixes (Q3).}
    \label{fig:teaser}
\end{figure}

\section{Introduction}
\label{sec:introduction}

Recognizing UAV intent is important for low-altitude airspace surveillance and security. Beyond detecting and tracking an aircraft, mission-intent recognition identifies the purpose of its observed behavior. Distinguishing transit, reconnaissance, probing, strike, and escape helps to interpret an encounter as it develops. In practice, intent must often be inferred before a complete trajectory is available, motivating evaluation from partial observation histories.

Conventional approaches predominantly rely on temporal deep learning models operating on numerical radar trajectories. Temporal networks~\cite{teng2021tcn,niu2025blac}, radar-specific representations~\cite{yun2023radar}, and behavior-to-language models~\cite{li2024behavior} have demonstrated the value of motion patterns for intent recognition. However, as illustrated in Fig.~\ref{fig:teaser}, when intermittent radar returns result in incomplete motion histories, these radar-only methods struggle to obtain sufficient behavioral evidence, reducing recognition accuracy. Moreover, the same mission intent can be expressed through different motion patterns and geometric configurations, which we refer to as \emph{tactical realizations}. The generalization of existing trajectory-based methods to unseen tactical realizations under incomplete radar observations remains insufficiently characterized. Visual observations provide scene and appearance cues that complement incomplete radar histories, while radar-derived behavioral descriptors make motion relations explicit. Combining these representations motivates a structured multimodal approach, whose generalization to unseen tactical realizations can be assessed using held-out trajectory programs.

Recent advances in multimodal large language models (MLLMs) offer a new avenue for integrating heterogeneous perceptual evidence with task semantics. JTD-UAV~\cite{wang2025jtduav} combines thermal UAV tracking with language-based behavior and intent description, without incorporating radar observations. Lei et al.~\cite{lei2026towards} further propose a framework connecting multimodal perception with intent reasoning, although its effectiveness has yet to be systematically evaluated. These studies suggest that MLLMs have the potential to integrate heterogeneous perceptual information with task semantics.

Nevertheless, existing research lacks a systematic benchmark and a unified evaluation protocol for multimodal intent recognition, particularly under unseen tactical realizations and incomplete radar observations. Existing benchmarks, including Anti-UAV~\cite{jiang2023antiuav}, MMAUD~\cite{yuan2024mmaud}, and LAMBDA~\cite{zhou2026lambda}, provide multimodal data for UAV tracking, localization, and low-altitude sensing and communication, but do not support mission-intent recognition with annotated intent labels.

To address this gap, we introduce \textbf{AeroIntent}, a simulation-based benchmark comprising 990 synchronized multimodal sequences, five mission intents, 22 trajectory-generation programs, three simulated environments, and 15 UAV--payload configurations. The dataset records radar, RGB, infrared, depth, and UAV state data. To evaluate generalization to unseen tactical realizations, the benchmark adopts a \emph{program-disjoint} protocol: the test set excludes all scene--configuration instances associated with trajectory programs used in training or validation. We also introduce controlled radar-gap and early intent recognition protocols to examine model performance under missing measurements and progressively accumulating observational evidence. We further propose \textbf{AeroMIR}, a structured multimodal intent reasoner that integrates radar trajectories, derived behavioral features, selected visual evidence, and prior task knowledge for candidate-based UAV intent recognition. Under the program-disjoint protocol across three seeds, AeroMIR achieves a mean Macro-F1 of 69.84\%, outperforming the strongest trajectory-only baseline by 7.79 percentage points and the radar--RGB late-fusion baseline by 19.32 percentage points. With a radar gap covering 40\% of the recording duration,
AeroMIR exceeds the strongest baseline by 6.30 Macro-F1
percentage points. In a separate observation-prefix
evaluation at P75, the margin is 9.88 points.

The main contributions of our work are:
\begin{itemize}
    \item We introduce \textbf{AeroIntent}, a synchronized multimodal benchmark for UAV intent recognition with trajectory-generation metadata and a program-disjoint protocol, enabling systematic evaluation under unseen tactical realizations and incomplete radar observations.
    \item We propose \textbf{AeroMIR}, a structured multimodal intent reasoner that integrates radar trajectories, behavioral features, visual evidence, and task knowledge through candidate-based fine-tuning.
    \item Our experiments demonstrate the effectiveness and \mbox{robustness} of AeroMIR under unseen tactical realizations, controlled radar gaps, and early intent recognition, outperforming competitive baselines and highlighting the value of multimodal and behavior-aware intent recognition.
\end{itemize}

\section{Related Work}
\label{sec:related-work}

\subsection{UAV Intent Recognition}
Motion-based intent recognition includes LSTM prediction with an entropy decision tree~\cite{zhou2020entropy}, radar trajectory-frequency classification~\cite{yun2023radar}, BiGRU-attention~\cite{teng2021gru}, attention--TCN--BiGRU~\cite{teng2021tcn}, and hierarchical LSTM with data enhancement~\cite{zhao2024enhanced}. Formation-level STFN combines formation geometry and swarm-centroid trajectories~\cite{li2025formation}. Li \emph{et al.} translate flight attributes into language~\cite{li2024behavior}, while Zheng \emph{et al.} use looming cues for early non-cooperative encounter recognition~\cite{zheng2025bio}. GenRec studies recognition from partial video observations~\cite{weng2024genrec}; AeroIntent instead evaluates multimodal mission-intent recognition on held-out UAV trajectory programs.

\subsection{Multimodal Perception and Incomplete Observations}
Multimodal learning studies complementary representations, alignment, and fusion~\cite{baltrusaitis2019multimodal}. Anti-UAV supports RGB--thermal tracking~\cite{jiang2023antiuav}; MMAUD provides camera, radar, LiDAR, and audio data for detection and localization~\cite{yuan2024mmaud}; LAMBDA combines visual, motion, LiDAR, radar, and wireless observations~\cite{zhou2026lambda}. These resources do not provide a unified labeled mission-intent evaluation with held-out trajectory programs.

In driving-scene 3D perception, CenterFusion associates radar measurements with image detections~\cite{nabati2021centerfusion}, and CRN fuses camera--radar features~\cite{kim2023crn}. SMIL addresses severely missing modalities~\cite{ma2021smil}, while Lee \emph{et al.} learn missing-modality-aware prompts for visual recognition~\cite{lee2023missing}. Our radar-gap protocol instead removes contiguous intervals within radar histories while retaining clean RGB evidence.

\subsection{Multimodal and Time-Series Language Models}
BLIP-2 bridges frozen pretrained image encoders and language models~\cite{li2023blip2}; LLaVA develops visual instruction tuning~\cite{liu2023visual}; PaLM-E incorporates sensory inputs into embodied language modeling~\cite{driess2023palme}. For UAV understanding, JTD-UAV combines thermal tracking and language description~\cite{wang2025jtduav}, while Lei \emph{et al.} organize perception summaries and task knowledge for intent reasoning~\cite{lei2026towards}. AeroIntent provides complementary labeled evaluations of generalization and incomplete observations.

For temporal data, LLMTIME encodes numerical sequences as text~\cite{gruver2023llmtime}, and Time-LLM reprograms time-series patches for a frozen language model~\cite{jin2024timellm}. These forecasting studies motivate careful numerical interfaces. AeroMIR combines radar states, deterministic behavioral descriptors, visual evidence, and task knowledge for candidate-based intent recognition.

\begin{figure*}[t]
\centering
\includegraphics[width=\textwidth]{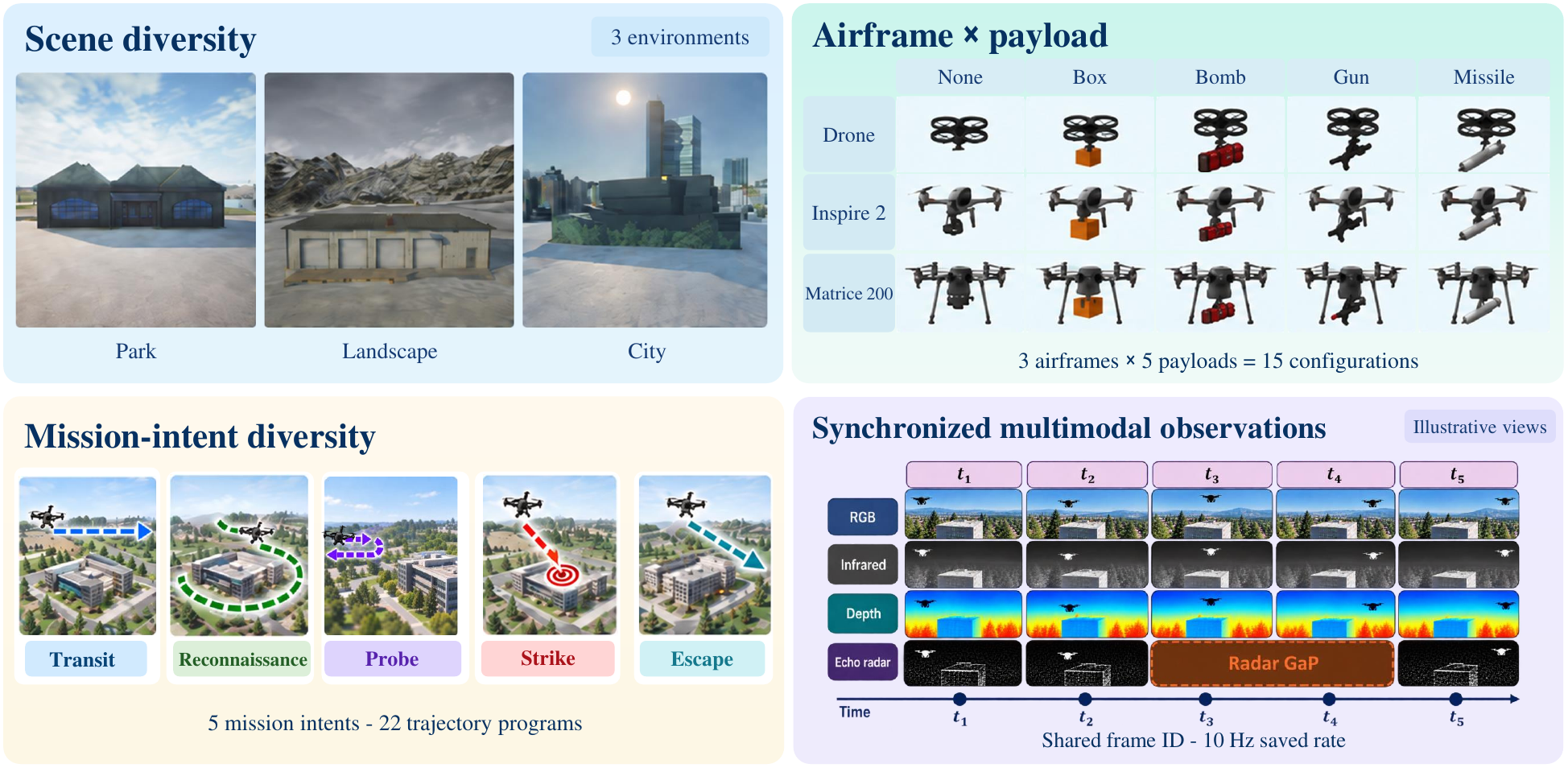}
\caption{\textbf{Overview of the AeroIntent benchmark.} The four panels summarize scene diversity, airframe-payload configurations, mission-intent diversity, and synchronized multimodal observations. The benchmark covers three environments, 15 UAV configurations, and five mission intents implemented through 22 trajectory programs. The temporal panel illustrates RGB, infrared, depth, and echo-radar observations at five representative timestamps. The highlighted radar-gap interval indicates missing radar observations while visual observations remain available. All streams are saved at 10~Hz with shared frame IDs. }

\label{fig:dataset}
\end{figure*}

\section{Benchmark: AeroIntent }
\label{sec:dataset}

\subsection{Task, Generation Factors, and Scale}
AeroIntent is generated in Unreal Engine~5 using Cosys-AirSim, an extension of AirSim~\cite{shah2018airsim,jansen2023cosysairsim}. Each observation is one complete, front-view flight sequence labeled by the objective of its generating program. The five classes describe \emph{mission intent}, rather than instantaneous maneuvers: transit traverses the monitored region; reconnaissance maintains an observation geometry; probe tests a boundary without committing to impact; strike converges on a target; and escape withdraws from the interaction.

As shown in Fig.~\ref{fig:dataset}, the AeroIntent benchmark crosses 22 trajectory programs with three scenes (park, mountainous landscape, and city) and 15 UAV configurations (three airframes, each with five payloads). The airframes are the AirSim drone, DJI Inspire~2, and DJI Matrice~200; payloads are none, box, bomb, gun, and missile. One sequence is recorded per combination, yielding $22\times3\times15=990$ sequences. Table~\ref{tab:split} gives class-level counts. All classes share the same scene and configuration factors.

\subsection{Sensing, Synchronization, and Annotations}
Collection saves observations at 10~Hz under simulation-pause control: sensor queries at a saved step refer to the same paused world state and share an integer \texttt{frame\_id}. The simulation clock is stored as \texttt{sim\_time\_ns} in nanoseconds. The configured 24-tick/s simulator advances with a nominal 2.4-tick sampling interval; actual timestamps, rather than an assumed constant interval, are used for velocity estimation. Radar--camera temporal correspondence follows the shared frame identifier.

The reference camera configuration uses $1080\times720$ images with a $90^\circ$ RGB field of view. Echo radar is configured for active sensing at 25~Hz, with 100,000 traces per measurement, a 500-m distance limit, up to four reflections, and azimuth/elevation limits of $[-90^\circ,90^\circ]$.  RGB, infrared, metric depth, Echo point clouds, IMU/GPS state, and frame annotations are retained. 

Simulator-provided target labels identify UAV returns in each Echo cloud, and their centroid defines the observed position. Consecutive timestamped centroids supply velocity where needed. Camera boxes come from simulator segmentation. Collection seeds are deterministically derived from the base seed, script, scene, UAV configuration, trajectory name, and repeat index. The benchmark metadata and configuration files preserve these choices.

\subsection{Program-Disjoint Evaluation Protocol}
\label{sec:dataset-protocol}
Structured distribution-shift benchmarks such as WILDS motivate evaluation beyond random sample splits~\cite{koh2021wilds}. We evaluate \emph{unseen tactical realizations} by holding out trajectory-generating programs, which specify motion patterns and their geometric configurations for a given intent. For each seed in $\{42,43,44\}$, we select one entire program per intent for testing. All 45 scene--configuration instances of each selected program are held out, giving 225 test sequences balanced at 45 per class. No test program appears in either training or validation: the split is strictly test-program-disjoint.

For transit, reconnaissance, probe, and strike, a second program is reserved for validation and all remaining programs are used for training. Escape has only two programs: after holding out one for testing, the remaining program is partitioned by UAV configuration into 36 training and nine validation sequences, keeping its three scene instances together. Thus, the escape training and validation sets share a program, while its test program remains unseen.  The seed controls program selection and training randomness. Each split contains 576/189/225 train/validation/test sequences.

\begin{table}[t]
\centering
\caption{Program counts and sequence allocation per seed. The test contains one unseen program per class.}
\label{tab:split}
\footnotesize
\setlength{\tabcolsep}{5pt}
\begin{tabular}{lrrrr}
\toprule
Intent & Programs & Train & Val & Test \\
\midrule
Transit & 6 & 180 & 45 & 45 \\
Reconnaissance & 6 & 180 & 45 & 45 \\
Probe & 3 & 45 & 45 & 45 \\
Strike & 5 & 135 & 45 & 45 \\
Escape & 2 & 36 & 9 & 45 \\
\midrule
Total & 22 & 576 & 189 & 225 \\
\bottomrule
\end{tabular}
\end{table}

\subsection{Additional Evaluation Protocols}
\label{sec:additional-protocols}

The \emph{radar-gap} protocol removes contiguous radar intervals at controlled severities; the \emph{early-intent} protocol retains observation prefixes. Both reuse each program-disjoint split's frozen test identities and trained checkpoints, isolating changes in available observations from changes in data partitioning. Section~\ref{sec:experiments} specifies the corresponding RGB-selection rules.

\begin{figure*}[t]
\centering
\includegraphics[width=\textwidth]{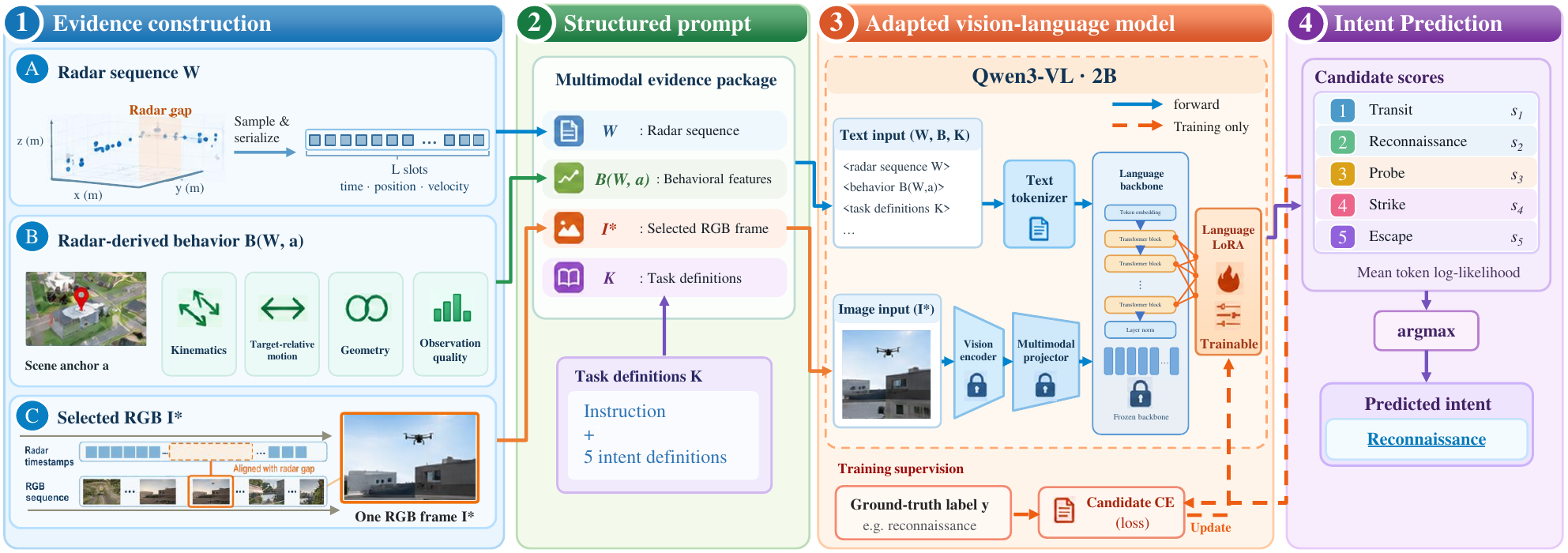}
\caption{\textbf{Overview of the AeroMIR.} Radar sequences, radar-derived behavioral features, and a gap-guided RGB frame are assembled with task definitions into a structured multimodal prompt. A vision-language model adapted with language LoRA scores five candidate intents using mean token log-likelihood, and the highest-scoring candidate is selected. Solid and dashed arrows indicate inference and training-only operations, respectively.}
\label{fig:method}
\end{figure*}

\section{Method: AeroMIR}
\label{sec:method}

\subsection{Structured Multimodal Evidence}
As shown in Fig.~\ref{fig:method}, AeroMIR connects numerical motion and visual context to a five-way intent decision. Given a radar window $W$, a scene-level target anchor $a$, and a synchronized RGB sequence, it forms a prompt
\begin{equation}
 P=\operatorname{Serialize}\big(W,B(W,a),I^*,K\big),
\end{equation}
where $B$ is a deterministic behavioral representation, $I^*$ is one selected image, and $K$ contains the candidate labels and task definitions. Let $\mathcal{Y}$ denote the five intent labels in Sec.~\ref{sec:dataset}. Behavior makes motion relations explicit within the prompt; RGB contributes scene and appearance information beyond radar. Both are interpreted by a shared vision-language backbone.

The anchor $a$ is a fixed scene reference in the same Echo-local metric coordinate system as the radar states. It is calibrated before evaluation by fitting Echo observations from dedicated reference-orbit calibration programs, with the world-to-Echo sign and offset checked against scripted geometry. These programs and their observations are disjoint from the train/validation/test splits; no held-out programs are used for calibration. Its horizontal coordinates are $(26.648,-0.082)$, $(21.670,-0.075)$, and $(26.594,0.085)$~m for park, landscape, and city. This scene geometry is assumed known at test time and shared across intents; neither the test sample's true intent nor its program-specific target center is used to select it.

\subsection{Compact Radar and Behavioral Representations}

Let $\{\tilde{\mathbf r}_i\}_{i=0}^{N-1}$ denote the
chronologically ordered valid target states.
We use $L$ temporal slots, where $L$ can be adjusted according to the available model context length and prompt budget; in this work, $L=30$, selecting
\begin{equation}
i_k=\operatorname{round}\!\left(\frac{k(N-1)}{L-1}\right),
\quad
\mathbf r_k=\tilde{\mathbf r}_{i_k},
\quad k=0,\ldots,L-1.
\end{equation}
This fixed-length interface limits prompt size and also
serves the trajectory baselines. Each prompt state contains
relative time (s), Echo-local position (m), velocity (m/s),
and a target-point-ratio confidence proxy.
Relative time is serialized to six decimal places,
and position and velocity to three.
When fewer than $L$ valid states are available, sampling
indices are repeated. With no valid target return,
slots are zero and marked invalid. Interior missing
intervals remain visible through timestamp/frame gaps,
rather than inserted zero positions.

The behavior processor summarizes four groups:
kinematics (duration and speed),
target-relative motion (range trend, radial velocity,
and approach/retreat fractions),
geometry (path efficiency, angular coverage, radius
stability, hovering, terminal linearity, and altitude change),
and observation quality (gaps, jumps, and confidence).
These deterministic transforms of radar observations
provide representation augmentation.
The computation is illustrated below using three
representative descriptors: radial velocity, path
efficiency, and radius stability.

For $k=0,\ldots,L-1$, let $\mathbf p_{k,xy}$ and
$\mathbf v_{k,xy}$ denote horizontal position and velocity,
respectively, and let $\mathbf a_{xy}$ be the target-region
anchor. Define
$\rho_k=\|\mathbf p_{k,xy}-\mathbf a_{xy}\|_2$.
Radial velocity is
\begin{equation}
v_k^{\mathrm{rad}}
=
\frac{
(\mathbf p_{k,xy}-\mathbf a_{xy})^\top
\mathbf v_{k,xy}
}{
\max(\rho_k,\epsilon_\rho)
},
\qquad
\epsilon_\rho=10^{-6}\,\mathrm{m}.
\end{equation}
Negative values indicate approach; the exported terminal
descriptor averages the last three slots.

Define the horizontal path length and displacement as
\begin{equation}
\begin{aligned}
S &= \sum_{k=1}^{L-1}
\|\mathbf p_{k,xy}-\mathbf p_{k-1,xy}\|_2,\\
D &= \|\mathbf p_{L-1,xy}-\mathbf p_{0,xy}\|_2.
\end{aligned}
\end{equation}
Let $\bar{\mathbf p}_{xy}$ be the mean horizontal position,
$r_k=\|\mathbf p_{k,xy}-\bar{\mathbf p}_{xy}\|_2$ the
centroid-relative radius, and $\bar r$ and $\sigma_r$ its
mean and population standard deviation.
Path efficiency and radius stability are
\begin{equation}
\begin{aligned}
\eta &=
\begin{cases}
C(D/S), & S\ge\epsilon,\\
0, & S<\epsilon,
\end{cases}\\
s_r &= C\!\left(
1-\frac{\sigma_r}{\max(\bar r,\epsilon)}
\right),
\end{aligned}
\end{equation}
where $C(u)=\min(1,\max(0,u))$ and
$\epsilon=10^{-9}\,\mathrm{m}$.
Means and population standard deviations include repeated
slots. These features describe target-relative motion,
path directness, and centroid-relative radius consistency.
Detailed definitions and computation procedures for the
complete feature set are provided in the accompanying
anonymous repository.
\subsection{Temporally Aligned Visual Evidence}
The standard selector prioritizes intervals following a valid radar hit with either a large inter-return gap (more than 20 frames) or missing target returns. It selects the RGB frame at the interval midpoint; recording-midpoint frame provides fallbacks. Only the selected image and its temporal relation to the radar window enter the prompt. Frame-choice sensitivity is evaluated in Sec.~\ref{sec:main-results}.

\subsection{Candidate-Based Adaptation and Prediction}
AeroMIR uses Qwen3-VL-2B-Instruct~\cite{qwen3vl} with QLoRA~\cite{dettmers2023qlora}. The base model is NF4 4-bit double-quantized. The vision encoder, multimodal projector, and base language weights remain frozen; rank-16 LoRA adapters~\cite{hu2022lora} (scaling 32, dropout 0.05) update language-side \texttt{q/k/v/o} and \texttt{gate/up/down} projections, totaling 17.43M trainable parameters. Images are capped at 262,144 pixels and the token sequence at 8192 tokens.

For candidate $c\in\mathcal{Y}$, let $\mathbf z_c$ be its tokenized assistant completion (label and chat-template termination tokens). Its score is
\begin{equation}
 s_c(P)=\frac{1}{|\mathbf z_c|}\sum_{u=1}^{|\mathbf z_c|}
 \log p_{\theta'}(z_{c,u}\mid P,z_{c,<u}).
\end{equation}
We optimize cross-entropy over these five sequence scores,
\begin{equation}
 \mathcal L_{\mathrm{cand}}=-\log\frac{\exp(s_y)}{\sum_{c\in\mathcal Y}\exp(s_c)},
 \qquad \hat y=\arg\max_{c\in\mathcal Y}s_c.
\end{equation}
Training and inference use the same candidate scores, which rank intents rather than calibrating risk. The supplied prompt template combines fixed task definitions and observed evidence.

An abridged prompt template is shown below; placeholders are replaced with sample-specific observations.

\begin{quote}
\footnotesize
Classify the UAV intent using the following evidence.

\smallskip
Intent definitions:
transit---passage with weak target interaction;
reconnaissance---orbiting, loitering, or repeated observation;
probe---tentative approach or approach-then-retreat;
strike---committed closing toward the target;
escape---evasive or high-speed departure.

\smallskip
Visible evidence JSON:
\begin{verbatim}
{
  "echo_trajectory": "<30 radar states>",
  "feature_primitive_prior": "<behavior cues>",
  "selected_rgb_frames": "<frame metadata>",
  "target_region": "<target coordinates>"
}
\end{verbatim}
\end{quote}

\section{Experiments}
\label{sec:experiments}

\subsection{Experimental Setup}
We evaluate three questions on the frozen tests in Sec.~\ref{sec:dataset-protocol}: intent recognition under unseen tactical realizations, with contiguous radar gaps, and early intent recognition. We report Accuracy and Macro-F1 as percentages, with population standard deviation across seeds 42/43/44. 

\noindent\textbf{Baselines.}
Six trajectory models use the same 30-slot $[x,y,z,v_x,v_y,v_z]$ input: MLP, LSTM, GRU, Attention-TCN-BiGRU~\cite{teng2021tcn}, BLAC cross-attention BiLSTM~\cite{niu2025blac}, and Behavior-to-Language~\cite{li2024behavior}. The latter three are benchmark adaptations of the cited designs. A radar--RGB late-fusion control combines a two-layer bidirectional GRU with a compact, from-scratch CNN and an MLP classifier, using the same selected RGB frame as AeroMIR. The Lei-style zero-shot reference follows the prompt organization in~\cite{lei2026towards}, replacing its backbone with Qwen3-VL-2B-Instruct and supplying ground-truth payload in place of YOLO output. These comparisons cover motion encoders, conventional multimodal fusion, and a language-based architectural reference.

AeroMIR additionally receives timestamps, frame-gap/validity information, and confidence in its structured evidence; the six-dimensional trajectory baselines do not receive separate time or mask channels. The comparisons therefore assess the complete input representations and models, rather than isolating the effect of multimodal fusion under identical temporal metadata.

\noindent\textbf{Training and evaluation.}
Classical models use train-only feature standardization, class-balanced loss, up to 80 epochs, and validation early stopping with patience 12. Late fusion uses $160\times160$ RGB, batch size 24, learning rate $10^{-3}$, weight decay $10^{-4}$, and dropout 0.2. AeroMIR uses AdamW, learning rate $2\times10^{-4}$, zero weight decay, 3\% warmup, batch size one, and eight-step gradient accumulation. It trains for 20 epochs with the candidate objective; the checkpoint with highest validation Macro-F1 among epochs 10--20 is selected, breaking ties by Accuracy and then later epoch. Gap and Early require inference only, without condition-specific tuning.

\subsection{Generalization to Unseen Tactical Realizations}
\label{sec:main-results}
\begin{figure}[t]
    \centering
    \includegraphics[width=\columnwidth]
    {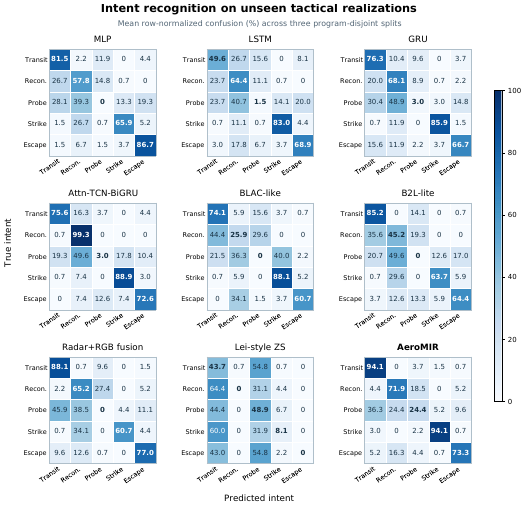}
    \caption{\textbf{Intent recognition under unseen tactical
    realizations.} Mean row-normalized confusion matrices across three
    program-disjoint splits for eight representative baselines and
    AeroMIR. Rows denote ground-truth intents and columns denote
    predicted intents. }
    \label{fig:confusion-comparison}
\end{figure}

Table~\ref{tab:main} reports the main results. AeroMIR achieves $71.56\%$ Accuracy and $69.84\%$ Macro-F1. Attention-TCN-BiGRU is the strongest trajectory baseline, at 67.85\% Accuracy and 62.05\% Macro-F1. AeroMIR improves these means by 3.71 and 7.79 points, respectively, and exceeds radar--RGB late fusion by 19.32 Macro-F1 points. Figure~\ref{fig:confusion-comparison} further resolves the recognition patterns of individual intent classes.

\begin{table}[t]
\centering
\caption{Intent recognition under unseen tactical realizations (\%, three-seed mean $\pm$ population standard deviation). R: radar; V: RGB; B: radar-derived behavior. ZS uses no fine-tuning and receives ground-truth payload.}
\label{tab:main}
\footnotesize
\setlength{\tabcolsep}{3pt}
\begin{tabular}{llcc}
\toprule
Method & Input & Accuracy & Macro-F1 \\
\midrule
MLP & R & $58.37\pm9.27$ & $53.60\pm8.58$ \\
LSTM & R & $53.48\pm11.34$ & $50.84\pm9.12$ \\
GRU & R & $60.00\pm4.04$ & $55.42\pm3.06$ \\
Attn-TCN-BiGRU~\cite{teng2021tcn} & R & $67.85\pm5.10$ & $62.05\pm5.95$ \\
BLAC-like~\cite{niu2025blac} & R & $49.78\pm8.20$ & $45.96\pm5.66$ \\
B2L-lite~\cite{li2024behavior} & R & $51.70\pm2.30$ & $48.08\pm4.25$ \\
Radar+RGB fusion & R+V & $58.22\pm10.08$ & $50.52\pm10.87$ \\
Lei-style ZS~\cite{lei2026towards} & R+V+GT & $20.15\pm1.11$ & $13.64\pm0.77$ \\
\rowcolor{aerolight}
\textbf{AeroMIR} & \textbf{R+B+V} &
\textbf{$71.56\pm5.16$} &
\textbf{$69.84\pm6.35$} \\
\bottomrule
\end{tabular}
\end{table}

The gains in transit, probe, and strike come at the cost of lower reconnaissance recall (71.9\% versus 99.3\% for Attention-TCN-BiGRU). AeroMIR improves transit recall from 75.56\% to 94.07\% and probe recall from 2.96\% to 24.44\% relative to Attention-TCN-BiGRU, while strike recall rises from 88.89\% to 94.07\%. 

The Lei-style zero-shot reference reaches 13.64\% Macro-F1 with the same backbone and ground-truth payload. Together with the supervised comparisons, this shows that access to a general-purpose MLLM and perception summaries alone does not solve the benchmark. The full improvement combines structured representation, pretraining, and supervised candidate adaptation.

\begin{figure}[t]
    \centering
    \includegraphics[width=\columnwidth]
    {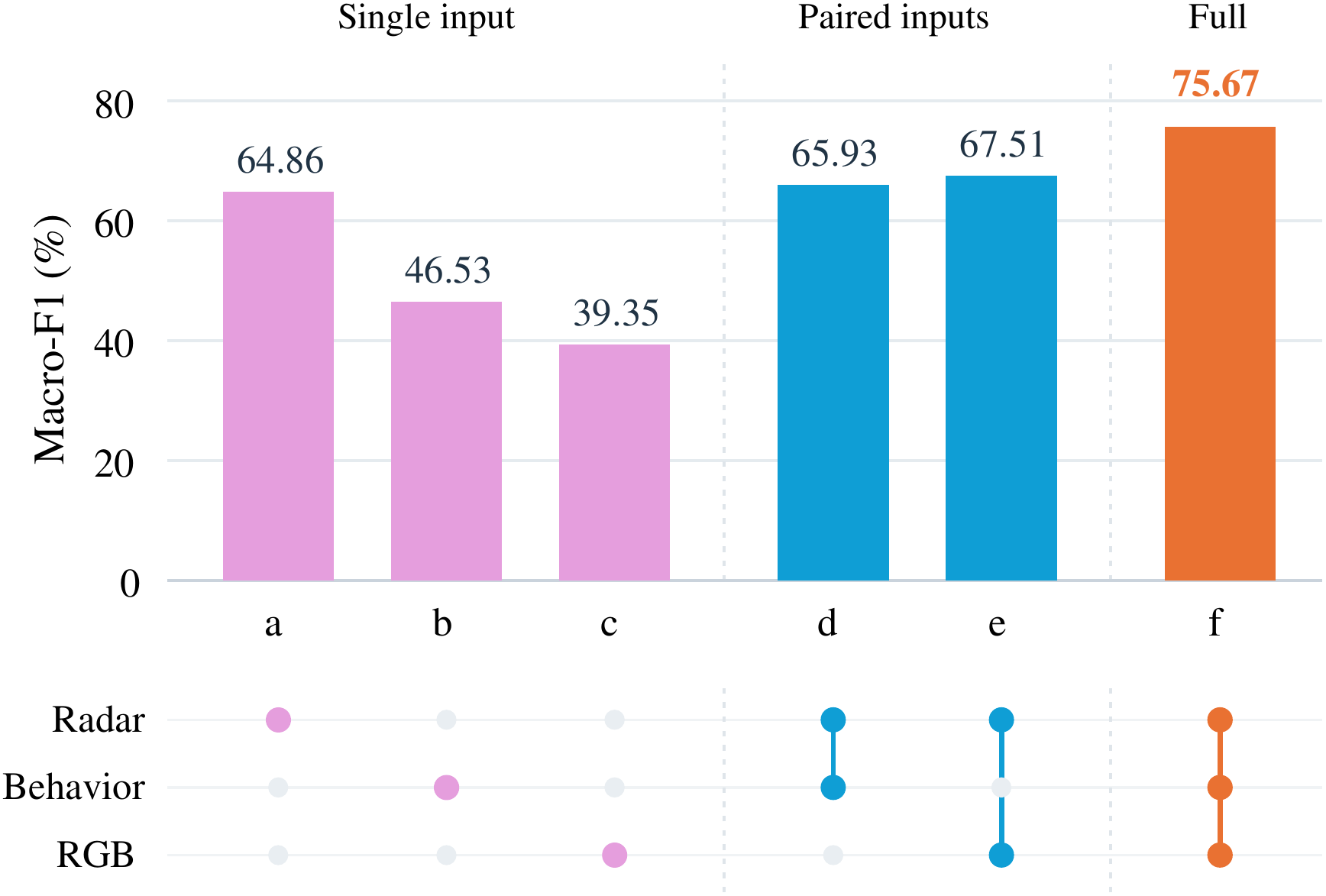}
    \caption{\textbf{Contribution of each evidence source.}
    Candidate-CE ablation on the split generated with seed 42. Bars report Macro-F1, and filled circles indicate the included representations. a--f correspond to R, B, V, R+B, R+V, and R+B+V, respectively, where R denotes radar trajectories, B denotes radar-derived behavioral descriptors, and V denotes RGB evidence.}
    \label{fig:evidence-ablation}
\end{figure}

\noindent\textbf{Ablation.}
On the split generated with seed 42, we train six candidate-CE variants using the same backbone, optimization settings, and frame-selection rule. a, b, and c use
radar (R), derived behavior (B), and RGB (V) individually, whereas
d, e, and f use R+B, R+V, and R+B+V, respectively. Their Macro-F1
values are 64.86, 46.53, 39.35, 65.93, 67.51, and $75.67\%$,
respectively, as shown in Fig.~\ref{fig:evidence-ablation}.
Adding behavior to radar gives 1.07 points, while adding it when RGB
is present gives 8.16 points. Adding RGB gives 2.65 points over radar
alone and 9.74 points over radar plus behavior. Thus, on this matched
split, the largest benefit occurs when explicit motion descriptors
and visual evidence are interpreted together. Neither behavior-only
nor RGB-only reproduces the full model's result.

\noindent\textbf{Frame-choice sensitivity.}
We compare gap-guided selection with random, midpoint, terminal, and highest-visibility frame selection on the complete program-disjoint test sets. Each of the three seeds (42/43/44) includes 225 test sequences, with 45 sequences per intent class. We keep the non-image evidence fixed, varying only the selected RGB frame. We report the unweighted mean of the three seed-wise Macro-F1: 69.84/68.10/68.67/68.46/67.95\%, respectively. The results show that the gap-guided rule provides a modest advantage over alternative frame-selection strategies, although the improvement is limited. This suggests that temporal frame selection is beneficial, but should be regarded as a practical evidence-selection choice rather than a major independent source of performance gains.

\begin{figure}[t]
\centering
\includegraphics[width=\columnwidth]{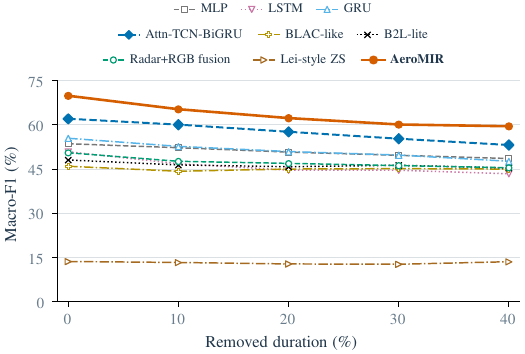}
\caption{\textbf{Controlled radar-gap recognition.}
Three-seed mean Macro-F1 for eight baselines and AeroMIR
under contiguous radar deletions covering 0--40\% of
the recording duration. The remaining radar observations
are resampled to 30 slots, while RGB observations remain
uncorrupted. AeroMIR and Radar--RGB Late Fusion apply
gap-guided RGB selection to the remaining radar observations;
the Lei-style zero-shot reference always uses the first
RGB frame. Gap0 matches the main experiment.}
\label{fig:gap}
\end{figure}

\subsection{Controlled Radar-Gap Recognition}
\label{sec:gap}

We evaluate recognition under contiguous radar outages using the same
program-disjoint test splits as the main experiment. For each sample,
we remove one contiguous interval from the original Echo sequence,
resample the remaining target states to 30 slots, and recompute
velocities and behavioral descriptors. Gap levels 0/10/20/30/40 specify
the removed interval's duration as a percentage of the full recording.
A sample-specific random anchor, fixed for reproducibility, defines
nested intervals across gap levels. All methods share the same radar
deletions, while RGB observations remain uncorrupted.
At each nonzero gap level, AeroMIR and Radar--RGB Late Fusion
reselect their RGB frames using the same gap-guided rule
applied to the remaining radar observations. The Lei-style
zero-shot reference consistently uses the first RGB frame
of the full recording. All models remain frozen throughout the evaluation.

As shown in Fig.~\ref{fig:gap}, AeroMIR achieves the highest mean
Macro-F1 across all five gap levels. With an interval covering 40\%
of the recording duration removed, it achieves 59.50\% Macro-F1,
exceeding the strongest trajectory baseline, Attention-TCN-BiGRU,
by 6.30 percentage points (53.20\%). It also outperforms radar--RGB
late fusion by 14.08 points (45.42\%), showing that the benefit extends
beyond simply providing both sensing modalities. Across the tested
gap range, AeroMIR achieves a normalized Macro-F1 AUC of 63.08,
compared with 57.67 for Attention-TCN-BiGRU
(Table~\ref{tab:temporal}). Its mean Macro-F1 decreases from 69.84\% at Gap0 to 59.50\% at Gap40 and remains the highest among the compared methods at all tested gap levels.
These results support the effectiveness of structured multimodal
evidence integration for intent recognition when contiguous portions
of the radar history are unavailable.

\begin{figure}[t]
\centering
\includegraphics[width=\columnwidth]{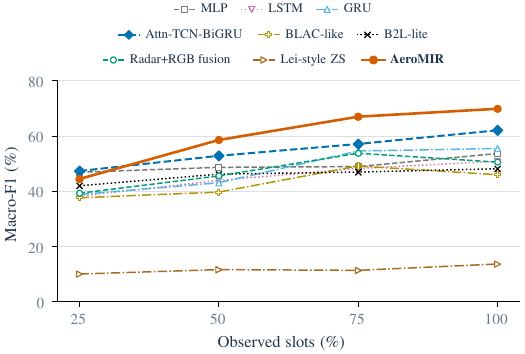}
\caption{\textbf{Early intent recognition.}
Three-seed mean Macro-F1 at 25\%, 50\%, 75\%, and 100\%
observation-slot coverage for eight baselines and AeroMIR.
Partial radar prefixes are resampled to 30 slots.
AeroMIR and Radar--RGB Late Fusion apply
gap-guided RGB selection within the observed prefix.
The Lei-style zero-shot reference uses the first RGB frame of the full recording at every prefix.}
\label{fig:early}
\end{figure}

\subsection{Early Recognition}
\label{sec:early}

We next evaluate how recognition performance evolves as observations
accumulate on the held-out programs.
Figure~\ref{fig:early} reports results at 25\%, 50\%, 75\%, and 100\%
coverage of the ordered 30-slot radar history, corresponding to
prefixes of 8, 15, 23, and 30 slots.
Each partial prefix is resampled to 30 slots, after which velocities
and behavioral descriptors are recomputed. For RGB selection, AeroMIR and Radar-RGB Late Fusion apply
gap-guided RGB selection using only radar observations and RGB
frames available up to the corresponding prefix endpoint. The Lei-style zero-shot reference consistently uses the first RGB frame of each recording. Coverage is defined over observation slots rather than elapsed
flight time. Each method uses its frozen full-data checkpoint throughout the evaluation.

AeroMIR demonstrates strong early intent recognition once
sufficient partial evidence becomes available. At P25, it
achieves 44.42\% Macro-F1, trailing Attention-TCN-BiGRU by
2.90 points. However, AeroMIR leads all compared
methods from P50 onward, reaching 58.54\% Macro-F1 at P50
and 66.97\% at P75, with gains of 5.72 and 9.88 percentage
points over Attention-TCN-BiGRU, respectively.
Notably, with only 75\% of the radar observation slots,
AeroMIR exceeds the strongest baseline's full-observation
Macro-F1 by 4.92 points and falls just 2.87 points below
its own full-observation performance. These results show
that AeroMIR can recognize task-level intent effectively
before the complete trajectory is observed.
Table~\ref{tab:temporal} summarizes performance using
normalized trapezoidal AUC over P25/P50/P75/P100.
AeroMIR achieves an Early AUC of 60.88, outperforming
Attention-TCN-BiGRU by 6.01 points and supporting its
overall advantage under partial observations.

\begin{table}[t]
\centering
\caption{Temporal evidence summary (mean Macro-F1, \%). AUC is the trapezoidal area normalized by the tested interval: gap0--40 or P25--100.}
\label{tab:temporal}
\footnotesize
\setlength{\tabcolsep}{4pt}
\begin{tabular}{lrrr}
\toprule
Method & Gap40 & Gap AUC & Early AUC \\
\midrule
MLP & 48.58 & 50.94 & 49.25 \\
LSTM & 43.39 & 45.78 & 45.56 \\
GRU & 47.63 & 51.21 & 48.22 \\
Attn-TCN-BiGRU~\cite{teng2021tcn} & 53.20 & 57.67 & 54.87 \\
BLAC-like~\cite{niu2025blac} & 44.98 & 44.96 & 43.54 \\
B2L-lite~\cite{li2024behavior} & 45.35 & 46.30 & 46.02 \\
Radar+RGB fusion & 45.42 & 47.16 & 48.06 \\
Lei-style ZS~\cite{lei2026towards} & 13.57 & 13.13 & 11.60 \\
\rowcolor{aerolight}
\textbf{AeroMIR} & \textbf{59.50} & \textbf{63.08} & \textbf{60.88} \\
\bottomrule
\end{tabular}
\end{table}

\section{Conclusion}
\label{sec:conclusion}
We introduce AeroIntent, a synchronized multimodal UAV mission-intent benchmark with program-disjoint, radar-gap, and observation-prefix protocols. AeroMIR combines radar trajectories, behavioral descriptors, visual evidence, and task knowledge through candidate-based prediction, achieving the highest mean Macro-F1 in the main comparison. It also achieves higher mean Macro-F1 under the tested radar-gap conditions and at observation prefixes of P50 and above, while trailing the strongest trajectory baseline at P25, highlighting the value of behavior-aware reasoning and complementary visual evidence.

Future work will incorporate AeroIntent's infrared, depth, and UAV-state modalities through reliability-aware fusion. The benchmark and method provide a reproducible basis for intent recognition under changing observation availability.

\balance
\bibliographystyle{IEEEtran}
\bibliography{main}
\end{document}